\documentclass[letterpaper, 10 pt, conference]{ieeeconf}  % Comment this line out if you need a4paper

\IEEEoverridecommandlockouts                              % This command is only needed if 
\usepackage{cite}
\usepackage{amsmath,amssymb,amsfonts}
\usepackage{algorithmic}
\usepackage{graphicx}
\usepackage{textcomp}
\usepackage{xcolor}
\usepackage[hidelinks]{hyperref}
\usepackage{subcaption}
\usepackage{booktabs}
\usepackage{multirow}
\usepackage{array}
\usepackage{mathrsfs}

\usepackage{orcidlink}
\usepackage{pifont}

\title{\LARGE \bf
Eye on the Target: Eye Tracking Meets Rodent Tracking
}

\author{
  Emil Mededovic \orcidlink{0009-0002-6801-4890}$^{\,1\,}$, 
  Yuli Wu \orcidlink{0000-0002-6216-4911}$^{\,1\,}$, 
  Henning Konermann \orcidlink{0009-0006-5736-5803}$^{\,1\,}$, \\
  Marcin Kopaczka \orcidlink{0000-0002-3695-4696}$^{\,1\,}$, 
  Mareike Schulz$^{\,2\,}$, 
  René Tolba$^{\,2\,}$, 
  Johannes Stegmaier \orcidlink{0000-0003-4072-3759}$^{\,1\,}$% <-this % stops a space
\thanks{This work was funded by the German Research Foundation DFG with the grants STE2802/4-1 (EM), TO 542/5-2 (RT), TO 542/6-2 (RT) and TO 542/9-1 (RT) as part of the DFG FOR 2591 Severity Assessment in Animal-Based Research. Furthermore, this work was funded by the German Research Foundation DFG with grant DFG RTG 2610 InnoRetVision (HK, YW).}%
\thanks{$^{1}$EM, YW, HK, MK, JS are with the Institute of Imaging and Computer Vision, RWTH Aachen University, Aachen, Germany. E-mail: \texttt{{emil.mededovic@lfb.rwth-aachen.de}}}%
\thanks{$^{2}$MS, RT are with the Institute for Laboratory Animal Science, RWTH Aachen University, Germany.}}

\begin{document}

\maketitle
\thispagestyle{empty}
\pagestyle{empty}

\begin{abstract}
Analyzing animal behavior from video recordings is crucial for scientific research, yet manual annotation remains labor-intensive and prone to subjectivity. Efficient segmentation methods are needed to automate this process while maintaining high accuracy. In this work, we propose a novel pipeline that utilizes eye-tracking data from Aria glasses to generate prompt points, which are then used to produce segmentation masks via a fast zero-shot segmentation model. Additionally, we apply post-processing to refine the prompts, leading to improved segmentation quality. Through our approach, we demonstrate that combining eye-tracking-based annotation with smart prompt refinement can enhance segmentation accuracy, achieving an improvement of 70.6\% from 38.8 to 66.2 in the Jaccard Index for segmentation results in the rats dataset.
% \newline

% \indent \textit{Clinical relevance}— This is a brief additional statement on why a this might be of interest to practicing clinicians. Example: This establishes the anesthetic efficacy of 10\% intraosseous injections with epinephrine to positively influence cardiovascular function.
\end{abstract}

% \begin{IEEEkeywords}
% Animal Experiment, Eye Tracking, Annotation 
% \end{IEEEkeywords}

\section{Introduction}
Laboratory animal experiments are essential for biomedical research, particularly in drug discovery \cite{sharpless2006mighty}. Among them, rodents are indispensable due to their genetic similarity to humans, small size and ease of maintenance \cite{bryda2013mighty}. They have been crucial in investigating genetic predisposition to obesity \cite{zhang1994positional} and remain vital in the evaluation of antiviral treatments, such as SARS-CoV-2 therapies \cite{bao2020pathogenicity}, underscoring their important role in the advancement of medical science. 

Analyzing animal experiments is critical for extracting meaningful insights, yet traditional human observation remains time-consuming, tedious, and often subjective \cite{gomez2014big}. This limitation affects the reproducibility and scalability of behavioral studies \cite{rosenthal1963effect, crabbe1999genetics, kafkafi2018reproducibility}. To address these challenges, automated video analysis technologies have emerged to streamline behavioral studies across various laboratory animals, including rodents \cite{pereira2022sleap, lauer2022multi}. Segmentation masks play a key role in this automation, enabling robust object identification and facilitating further downstream analysis. Recent advances in off-the-shelf zero-shot models have made segmentation more efficient and adaptable across various domains \cite{kirillov2023segment, ravi2024sam}. These methods support a range of applications, such as tracking animal movement to study behavior in relation to neural mechanisms \cite{gomez2014big}, but also more general tasks like posture estimation and interaction analysis.

Despite these advancements, all such methods rely on labeled training data, requiring extensive human annotation. To streamline this process and showcase the potential for future automated animal behavior analysis, we explore novel annotation approaches inspired by gaze and eye tracking. Prior studies have demonstrated the effectiveness of gaze-based annotation in medical image segmentation, employing gaze as weak labels for supervised learning \cite{zhong2024weakly} and for directly collecting segmentation masks through gaze interaction \cite{wang2023gazesam, khaertdinova2024gaze}.

In this work, we propose a novel pipeline that utilizes eye-tracking from Aria glasses \cite{engel2023project} to generate prompt points, which are then used to produce segmentation masks via the fast zero-shot segmentation model, EfficientSAM \cite{xiong2024efficientsam} similar to the work \cite{gunther2020bionic}. EfficientSAM \cite{xiong2024efficientsam} is a lightweight SAM model trained using masked image pretraining. Additionally, we explore easy-to-use post-processing approaches that do not require additional training to refine and enhance the segmentation quality. To evaluate the feasibility and effectiveness of this approach in real-world scenarios, we conduct a user study with nine human participants.

\section{Methods}
In the following, we describe the data used and outline the pipeline from eye movement to prompt, leading to the segmentation mask. Additionally, we detail the post-processing techniques employed to refine the prompts.
\begin{figure*}[h!] 
\centering
\includegraphics[width=0.9\textwidth]{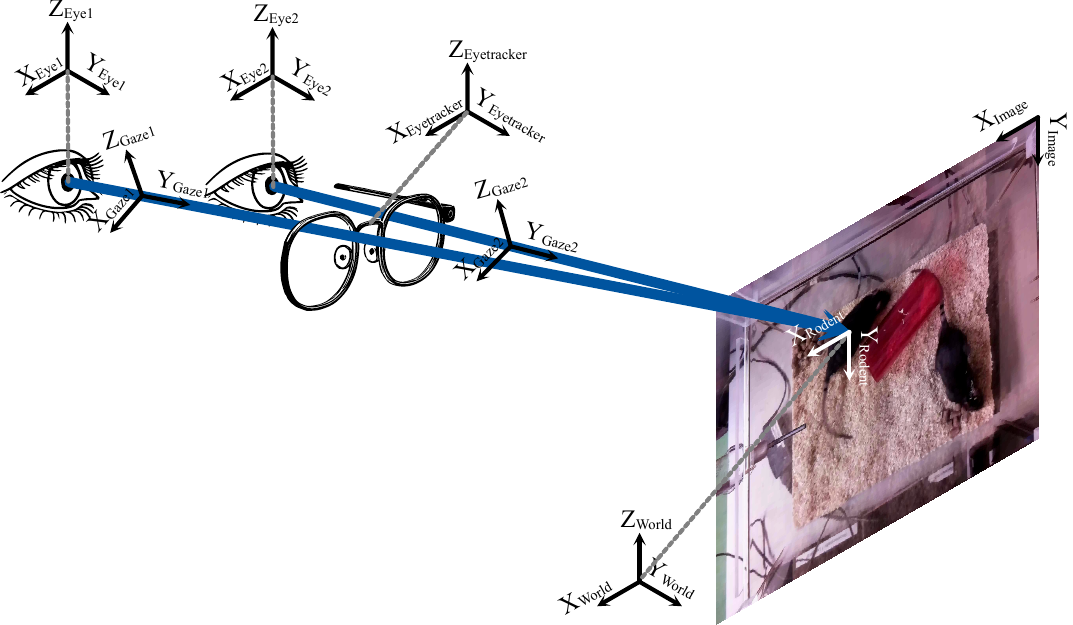} % Adjust width
\caption{Aria glasses \cite{engel2023project} integrate cameras to monitor eye movements and capture the user's field of view. The gaze estimation process consists of the following transformations: (1) \textbf{Eye to Eye Tracker} (\(\textbf{T}_{\mathrm{EET}}\)): Capturing eye movements via dedicated cameras. (2) \textbf{Eye Tracker to Gaze} (\(\textbf{T}_{\mathrm{ETG}}\)): Deriving gaze direction from eye tracker data. (3) \textbf{Gaze to World} (\(\textbf{T}_{\mathrm{GW}}\)): Mapping gaze coordinates onto the world coordinate system. (4) \textbf{World to Image} (\(\textbf{T}_{\mathrm{WI}}\)): Projecting world coordinates onto the image plane. The extracted gaze-based prompts serve as inputs for segmentation using EfficientSAM \cite{xiong2024efficientsam}, with optional post-processing to refine and enhance the segmentation quality.}
\label{fig:overview}
\end{figure*}
\subsection{Experiments}
\begin{figure*} 
\centering
\includegraphics[width=\textwidth]{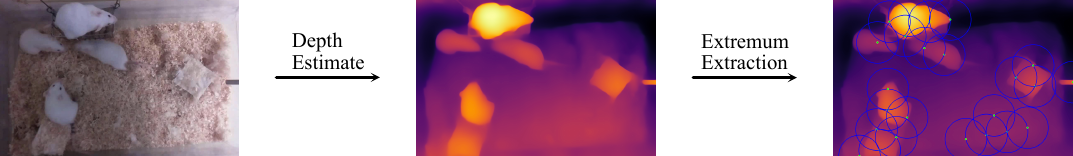} % Adjust width
\caption{For depth-aware refinement, we begin by computing the depth map using Depth-Anything v2 \cite{yang2024depth}. We then iteratively identify local maxima, ensuring that each selected peak is accompanied by a zone of exclusion to prevent overcrowding and the selection of adjacent, closely spaced peaks.}
\label{fig:mice}
\end{figure*}

The data consists of two datasets, each containing a single 30-second video recorded at 30 fps with a resolution of 1640×1232 pixels. The first video features two black rats (Fig. \ref{fig:overview}), while the second video includes four white mice (Fig. \ref{fig:mice}). We conducted experiments with nine participants, each tracking two rats and two mice across three runs.
\subsection{Overview Pipeline}
The Aria glasses \cite{engel2023project} are equipped with cameras that track eye movements and record the scene in front of the wearer as seen in Fig. \ref{fig:overview}. The system consists of multiple coordinate frames as depicted in Fig. \ref{fig:overview}. The transformation sequence follows a structured flow:

\begin{enumerate}
    \item \textbf{Eye to Eye Tracker} (\(\textbf{T}_{\mathrm{EET}}\)): The eye tracker cameras record eye movements.
    \item \textbf{Eye Tracker to Gaze} (\(\textbf{T}_{\mathrm{ETG}}\)): The system computes gaze direction from the eye tracker data.
    \item \textbf{Gaze to World} (\(\textbf{T}_{\mathrm{GW}}\)): Gaze coordinates are transformed into the world coordinate system.
    \item \textbf{World to Image} (\(\textbf{T}_{\mathrm{WI}}\)): Projection of world coordinates onto the image plane.
\end{enumerate}

Since these transformations are not known beforehand, Aria \cite{engel2023project} employs a trained model to estimate gaze positions in world coordinates. To map the rodent onto the image frame, an additional perspective transformation is necessary. This is achieved through edge-based detection \cite{canny1986computational} followed by thresholding, which identifies the borders of the rectangular polygon required for the transformation. The extracted gaze-based prompts are used as inputs for segmentation via EfficientSAM [17], with optional post-processing applied to further refine the prompts and subsequently enhance the segmentation quality.

\subsection{Local Exploratory Sampling}
Local Exploratory Sampling (LES) is used to refine prompt points when the segmentation result is unreliable. Instead of relying solely on a single prompt, LES explores multiple nearby candidate points and selects the best one based on segmentation quality.
Given an initial prompt $\mathbf{p} = (x_{\mathrm{p}}, y_{\mathrm{p}})$ and a segmentation mask $\mathcal{M} : \Omega \to \{0,1\}$, where $\Omega \subset \mathbb{N}^2$ represents the discrete image domain and $\mathcal{M}(x,y) = 1$ indicates a foreground pixel whereas $\mathcal{M}(x,y) = 0$ specifies a background pixel, the total segmentation mask size is computed as:

\begin{equation} \mathcal{S} = \sum_{(x,y) \in \Omega} \mathcal{M}(x,y), \end{equation}
where $\mathcal{S}$ represents the number of foreground pixels. If the computed mask size $\mathcal{S}$ deviates significantly from the expected average size, it is considered unreliable. The acceptable range for the segmentation size is given by:

\begin{equation}
    \mathcal{S}_{\min} = \mathbb{E}[\mathcal{S}] \cdot (1 - \alpha),
\end{equation}

\begin{equation}
    \mathcal{S}_{\max} = \mathbb{E}[\mathcal{S}] \cdot (1 + \alpha)
\end{equation}
Where $\alpha$ is the allowed relative mask size deviation. If the mask size falls outside the valid range:
\begin{equation}
    \mathcal{S} < \mathcal{S}_{\min} \quad \text{or} \quad \mathcal{S} > \mathcal{S}_{\max},
\end{equation}
LES generates $N$ candidate prompts within a local area with points
\begin{equation}
    \mathbf{p}_i = (x_p + \Delta x_i, y_p + \Delta y_i), \quad \Delta x_i, \Delta y_i \sim \mathcal{U}(-r, r),
\end{equation}
where $\mathcal{U}(-r, r)$ is a uniform distribution ensuring that sampled points remain within the local neighborhood. For each candidate prompt $\mathbf{p}_i$, a segmentation mask $\mathcal{M}_i$ is obtained, and its size $\mathcal{S}_i$ is computed. Instead of searching for the globally optimal prompt, the first valid prompt satisfying the size constraint:

\begin{equation}
    \mathcal{S}_{\min} \leq \mathcal{S}_i \leq \mathcal{S}_{\max}
\end{equation}

is selected immediately to minimize computation. If no valid segmentation is found within the sampled points, the original prompt $\mathbf{p}$ is retained. 
\subsection{Kalman Filter}
Another way to guide the prompt is to use a Kalman Filter \cite{kalman1960new}. If a prompt $\mathbf{p} = (x_p, y_p)$ produces a segmentation mask $\mathcal{M}$ whose size is given again by Eq. (1) and the criterion in Eq. (4) is met, then the Kalman Filter estimates a new prompt position. The Kalman state vector is
\begin{equation}
    \mathbf{x}_t = 
    \begin{bmatrix}
        x_t \\ y_t \\ v_{x,t} \\ v_{y,t}
    \end{bmatrix},
\end{equation}

where $(x_t, y_t)$ is the current prompt position, and $(v_{x,t}, v_{y,t})$ is the velocity. The predicted prompt is given by

\begin{equation}
    (x_{\text{KF}}, y_{\text{KF}}) = \mathbf{H} \hat{\mathbf{x}}_t,
\end{equation}
where \textbf{H} is the observation matrix and will be specified in the implementation details. This new prompt is used as input to the segmentation model:

\begin{equation}
    \mathcal{M}_{\text{KF}} = f(x_{\text{KF}}, y_{\text{KF}}),
\end{equation}

where $f(\cdot)$ represents the segmentation function. If $\mathcal{M}_{\text{KF}}$ satisfies the size constraints, the prompt is updated:

\begin{equation}
    \mathbf{p} = (x_{\text{KF}}, y_{\text{KF}}).
\end{equation}

The Kalman Filter is only updated when the initial segmentation mask size is already valid:

\begin{equation}
    \mathcal{S} \in [\mathcal{S}_{\min}, \mathcal{S}_{\max}].
\end{equation}

In that case, the observed prompt $\hat{\mathbf{x}}_t = (x_p, y_p)$ is used to correct the filter using the Kalman gain $\mathbf{K}_t$:

\begin{equation}
    \mathbf{x}_{t+1} = \hat{\mathbf{x}}_t + \mathbf{K}_t \hat{\mathbf{x}}_t - \mathbf{H} \hat{\mathbf{x}}_t.
\end{equation}

This ensures that the Kalman Filter is always based on reliable segmentation results and does not incorporate erroneous measurements.
\subsection{Depth Aware Refinement}
Depth Aware Refinement (DAR) improves segmentation accuracy by refining the prompt location using depth information. Instead of relying solely on a user-specified prompt, depth maxima are identified and used to relocate the prompt to a more suitable position. Depth maps are estimated using Depth-Anything v2 \cite{yang2024depth}, given an input image $\mathcal{I}$:

\begin{equation}
    \mathcal{D} = g(\mathcal{I}),
\end{equation}

where $g(\cdot)$ is the zero-shot depth estimation function. To identify regions of interest, we iteratively extracted local maxima from the depth map while ensuring that selected points are spatially distinct. At each iteration, the highest remaining depth value is chosen,

\begin{equation}
    \mathbf{m}_i = \arg\max_{\mathbf{x} \in \mathcal{D}} \mathcal{D}(\mathbf{x}),
\end{equation}

and added to the set of maxima where $\textbf{x}=(x, y)$. To prevent clustering, pixels within a radius $r$ around each selected maximum are suppressed,

\begin{equation}
    \mathcal{D}(\mathbf{x}) = -\infty, \quad \forall \mathbf{x} \text{ such that } ||\mathbf{x} - \mathbf{m}_i||_2 \leq r.
\end{equation}

This ensures that depth maxima are well-distributed across the image similar to \cite{mededovic}. Once depth maxima are extracted, we determine the closest maximum to the original prompt $\mathbf{p}$ and set it as the refined prompt:

\begin{equation}
    \mathbf{p}^* = \arg\min_{\mathbf{m}_i \in \mathbf{M}} ||\mathbf{m}_i - \mathbf{p}||_2.
\end{equation}

This ensures that the refined prompt is both spatially close to the original prompt and aligned with a prominent object region in the depth map. The refined prompt is then used for segmentation.
\subsection{Implementation Details}
The relative mask size deviation $\alpha = 0.5$, with the valid range being computed from the first frame using the average annotated masks.

A total of $N = 20$ alternative prompts are tested, with the area size varying according to the dataset. We choose $r = 50$ and $r=25$ for rats and mice respectively.

In Kalman filters, the state transition matrix $\mathbf{F}$ and measurment matrix $\mathbf{H}$ are set such that position and velocity are jointly estimated, while only the position is measureable:
\begin{equation}
    \mathbf{F} =
    \begin{bmatrix}
        1 & 0 & 1 & 0 \\
        0 & 1 & 0 & 1 \\
        0 & 0 & 1 & 0 \\
        0 & 0 & 0 & 1
    \end{bmatrix},\
    \mathbf{H} =
    \begin{bmatrix}
        1 & 0 & 0 & 0 \\
        0 & 1 & 0 & 0
    \end{bmatrix}.
\end{equation}

The Process noise covariance is $\mathbf{Q}=10^{-2}\cdot\mathbf{I}_4$, while measurement noise covariance is $\mathbf{R} = 10^{-1}\cdot\mathbf{I}_2$.

The number of maxima and exclusion radius are dataset-dependent for depth-aware refinement as seen in Table \ref{tab:depth_params}. We qualitatively ensured that at least one local maximum landed on the object of interest at each time step.
\begin{table}[h]
    \centering
    \normalsize
     \begin{tabular}{lcc}
        \hline
        Dataset & \#maxima & r \\ 
        \hline
        Rats  & 8  & 200  \\ 
        Mice  & 22 & 125  \\ 
        \hline
    \end{tabular}
    \caption{Depth-aware refinement parameters for different datasets.}
    \label{tab:depth_params}
\end{table}
\subsection{Evaluation Metrics}
The Jaccard Index (J) \cite{jaccard1901etude} and Dice Soerenson Coefficient (DSC) \cite{dice1945measures} are common metrics for evaluating segmentation overlap. Let \( A \) denote the predicted region and \( B \) the ground truth region. The Jaccard Index is defined as:

\begin{equation}
\mathrm{J}(A, B) = \frac{|A \cap B|}{|A \cup B|}    
\end{equation}

and the Dice Soerenson Coefficient as:

\begin{equation}
\mathrm{DSC}(A, B) = \frac{2|A \cap B|}{|A| + |B|}    
\end{equation}

Both evaluation metrics range from 0 to 1, where 1 indicates perfect agreement between the predicted (\( A \)) and ground truth (\( B \)) regions. For better readability, we scaled the scores from 0 to 100.
\section{Results}
\begin{table*}[!]
    \caption{The table presents the performance for video object segmentation across multiple participants (p1 to p9) for both rats and mice. Three different post-processing components—Local Exploratory Sampling, Kalman Filtering, and Depth Aware Refinement—are tested in isolation to assess their impact on segmentation quality. The evaluation metrics include the Jaccard Index (J) \cite{jaccard1901etude} and Dice Soerenson Coefficient (DSC) \cite{dice1945measures}, where higher values indicate better segmentation accuracy.  Boldface results indicate the best performance, while underlined results represent the second-best performance. For better readability, we scaled the scores from 0 to 100.
}
    \centering
    \small
    \begin{tabular}{cccccccc}
    \toprule
      \multirow{2}{7em}[-0.5em]{\centering Participants}   & \multirow{2}{8em}[-0.5em]{\centering Local Exploratory Sampling} & \multirow{2}{5.5em}[-0.5em]{\centering Kalman Filter} & \multirow{2}{5.5em}[-0.5em]{\centering Depth Aware Refinement}  & \multicolumn{2}{c}{Rats} &  \multicolumn{2}{c}{Mice} \\ \cmidrule(lr{0.4em}){5-6} \cmidrule(l{0.4em}r){7-8}
      & &  & & {\centering $\mathrm{J}$ $\uparrow$}  & {\centering $\mathrm{DSC}$ $\uparrow$} & {\centering $\mathrm{J}$ $\uparrow$}  & {\centering $\mathrm{DSC}$ $\uparrow$} \\\midrule
        \multirow{4}{*}{p1} & \ding{55}  & \ding{55} & \ding{55} & 24.3 & 28.4 & 31.0 & 33.9 \\
        & \ding{51} & \ding{55} & \ding{55} & \underline{38.9} & \underline{43.4} & \underline{37.7} & \underline{40.7} \\
        & \ding{55} & \ding{51} & \ding{55} & 28.9 & 33.0 & 36.0 & 39.0 \\
        & \ding{55} & \ding{55} & \ding{51} & \textbf{60.7 (+36.4)} & \textbf{67.9 (+39.5)} & \textbf{41.8 (+10.8)} & \textbf{45.6 (+11.7)} \\
        \midrule
        \multirow{4}{*}{p2} & \ding{55}  & \ding{55} & \ding{55} & 16.5 & 19.4 & 7.1 & 8.8 \\
        & \ding{51} & \ding{55} & \ding{55} & \underline{24.4} & \underline{27.5} & \underline{10.5} & \underline{12.4} \\
        & \ding{55} & \ding{51} & \ding{55} & 19.5 & 22.4 & 8.4 & 10.2 \\
        & \ding{55} & \ding{55} & \ding{51} & \textbf{61.2 (+44.7)} & \textbf{68.2 (+48.8)} & \textbf{26.7 (+19.6)} & \textbf{29.2 (+20.4)} \\
        \midrule
        \multirow{4}{*}{p3} & \ding{55}  & \ding{55} & \ding{55} & 63.3 & 71.2 & 55.3 & 60.0 \\
        & \ding{51} & \ding{55} & \ding{55} & \underline{70.0} & \underline{77.5} & \textbf{67.9 (+12.6)} & \textbf{73.2 (+13.2)} \\
        & \ding{55} & \ding{51} & \ding{55} & 64.3 & 71.8 & \underline{61.5} & \underline{66.5} \\
        & \ding{55} & \ding{55} & \ding{51} & \textbf{71.7 (+8.4)} & \textbf{80.4 (+9.2)} & 59.4 & 64.4 \\
        \midrule
        \multirow{4}{*}{p4} & \ding{55}  & \ding{55} & \ding{55} & 55.6 & 62.9 & 21.5 & 23.7 \\
        & \ding{51} & \ding{55} & \ding{55} & \underline{68.2} & \underline{75.5} & \underline{36.6} & \underline{39.6} \\
        & \ding{55} & \ding{51} & \ding{55} & 58.6 & 65.8 & 27.3 & 29.7 \\
        & \ding{55} & \ding{55} & \ding{51} & \textbf{70.3 (+14.7)} & \textbf{79.0 (+16.1)} & \textbf{46.2 (+24.7)} & \textbf{50.0 (+26.3)} \\
        \midrule
        \multirow{4}{*}{p5} & \ding{55}  & \ding{55} & \ding{55} & 67.7 & 76.3 & 47.4 & 51.6 \\
        & \ding{51} & \ding{55} & \ding{55} & \textbf{73.0 (+5.3)} & \textbf{81.2 (+4.9)} & \underline{61.4} & \underline{66.2} \\
        & \ding{55} & \ding{51} & \ding{55} & 69.4 & 77.8 & 54.0 & 58.3 \\
        & \ding{55} & \ding{55} & \ding{51} & \underline{71.9} & \underline{80.7} & \textbf{63.5 (+16.1)} & \textbf{68.7 (+17.1)} \\
        \midrule
        \multirow{4}{*}{p6} & \ding{55}  & \ding{55} & \ding{55} & 38.6 & 44.5 & 36.6 & 40.0 \\
        & \ding{51} & \ding{55} & \ding{55} & \underline{59.4} & \underline{65.9} & \textbf{51.1 (+14.5)} & \textbf{55.2 (15.2)} \\
        & \ding{55} & \ding{51} & \ding{55} & 43.6 & 49.4 & 41.5 & 45.1 \\
        & \ding{55} & \ding{55} & \ding{51} & \textbf{70.3 (+31.7)} & \textbf{78.7 (+34.2)} & \underline{44.5} & \underline{48.2} \\
        \midrule
        \multirow{4}{*}{p7} & \ding{55}  & \ding{55} & \ding{55} & 22.7 & 26.6 & 20.8 & 23.7 \\
        & \ding{51} & \ding{55} & \ding{55} & \underline{37.7} & \underline{41.9} & \underline{32.8} & \underline{36.1} \\
        & \ding{55} & \ding{51} & \ding{55} & 27.7 & 31.6 & 25.1 & 28.0 \\
        & \ding{55} & \ding{55} & \ding{51} & \textbf{63.9 (+41.2)} & \textbf{71.2 (+44.6)} & \textbf{38.9 (+18.1)} & \textbf{42.4 (+18.7)} \\
        \midrule
        \multirow{4}{*}{p8} & \ding{55}  & \ding{55} & \ding{55} & 40.9 & 46.4 & 24.5 & 26.9 \\
        & \ding{51} & \ding{55} & \ding{55} & \underline{60.8} & \underline{67.4} & \underline{37.0} & \underline{39.9} \\
        & \ding{55} & \ding{51} & \ding{55} & 47.1 & 52.7 & 31.3 & 34.0 \\
        & \ding{55} & \ding{55} & \ding{51} & \textbf{68.6 (+27.7)} & \textbf{77.1 (+30.7)} & \textbf{45.6 (+21.1)} & \textbf{50.1 (+23.2)} \\
        \midrule
        \multirow{4}{*}{p9} & \ding{55}  & \ding{55} & \ding{55} & 19.1 & 23.4 & 2.7 & 3.4 \\
        & \ding{51} & \ding{55} & \ding{55} & \underline{34.1} & \underline{39.0} & 5.5 & 6.3 \\
        & \ding{55} & \ding{51} & \ding{55} & 24.7 & 29.1 & \underline{5.7} & \underline{6.4} \\
        & \ding{55} & \ding{55} & \ding{51} & \textbf{56.8 (+37.7)} & \textbf{64.2 (+40.8)} & \textbf{26.1 (+23.4)} & \textbf{28.9 (+25.5)} \\\bottomrule \bottomrule
        \multirow{4}{*}{$\overline{p}$} & \ding{55} & \ding{55} & \ding{55} & 38.8 & 44.4 & 27.4 & 30.2 \\
        & \ding{51} & \ding{55} & \ding{55}  & \underline{51.8} & \underline{57.7} & \underline{37.8} & \underline{41.0} \\
        & \ding{55} & \ding{51} & \ding{55}  & 42.6 & 48.2 & 32.3 & 35.3 \\
        & \ding{55} & \ding{55} & \ding{51}  & \textbf{66.2 (+27.4)} & \textbf{74.2 (+29.8)} & \textbf{43.6 (+16.2)} & \textbf{47.5 (+17.3)} \\\bottomrule
    \end{tabular}
    \label{tab:result_raw}
\end{table*}

\begin{table*}[t]
    \caption{The table presents the performance for video object segmentation for both rats and mice. Three different post-processing components are tested in combination: local exploratory sampling, Kalman filtering, and depth aware refinement to assess their impact on segmentation quality. Evaluation metrics include the Jaccard Index (J) \cite{jaccard1901etude} and the Dice Soerenson Coefficient (DSC) \cite{dice1945measures}, where higher values indicate better segmentation accuracy. Boldface results indicate the best performance, while underlined results represent the second-best performance. For better readability, we scaled the scores from 0 to 100.}
    \centering
    \small
    \begin{tabular}{cccccccc}
    \toprule
      \multirow{2}{8.5em}[-0.5em]{\centering Participants}   & \multirow{2}{8em}[-0.5em]{\centering Local Exploratory Sampling} & \multirow{2}{5.5em}[-0.5em]{\centering Kalman Filter} & \multirow{2}{5.5em}[-0.5em]{\centering Depth Aware Refinement}  & \multicolumn{2}{c}{Rats} &  \multicolumn{2}{c}{Mice} \\ \cmidrule(lr{0.4em}){5-6} \cmidrule(l{0.4em}r){7-8}
      & &  & & {\centering $\mathrm{J}$ $\uparrow$}  & {\centering $\mathrm{DSC}$ $\uparrow$} & {\centering $\mathrm{J} \uparrow$}  & {\centering $\mathrm{DSC}$ $\uparrow$} \\\midrule
        \multirow{5}{*}{$\overline{p}$} & \ding{55} & \ding{55} & \ding{51}  & 66.2 & 74.2 & 43.6 & 47.5 \\
        & \ding{51} & \ding{51} & \ding{55} & 51.6 & 57.3 & 39.1 & 42.3 \\
        & \ding{51} & \ding{55} & \ding{51} & \textbf{68.4 (+2.2)} & \textbf{75.8 (+1.6)} & \underline{48.7} & \underline{52.7} \\
        & \ding{55} & \ding{51} & \ding{51} & 67.0 & 74.6 & 48.3 & 52.4 \\
        & \ding{51} & \ding{51} & \ding{51} & \underline{68.1} & \underline{75.3} & \textbf{50.9 (+7.3)} & \textbf{55.0 (+7.5)} \\\bottomrule
    \end{tabular}
    \label{tab:result}
\end{table*}
In the Table \ref{tab:result_raw}, we observe a significant performance difference between the results obtained on the rats and mice datasets. This discrepancy can be quantified in terms of the Jaccard Index ($\mathrm{J}$), where we achieve a score of 38.8 for rats data compared to 27.4 for mice, and the Dice Soerenson Coefficient ($\mathrm{DSC}$) is 44.4 for rats versus 30.2 for mice.

The primary reason for this disparity are the inherent differences in the datasets. The rats dataset features only two larger rats, making them easier to track visually, whereas the mice dataset contains four smaller and faster-moving mice, which complicates tracking and segmentation. The contrast in background conditions further influences performance: black rats are tracked against a white background, whereas white mice appear against a white background, making them harder to distinguish. Additionally, we observe a substantial performance variation across individual participants. For example, participant p5 achieves a Jaccard Index of 67.7 for rats and 47.4 for mice, whereas participant p9 performs significantly worse, with $\mathrm{J}$ = 19.1 for rats and an extremely low $\mathrm{J}$ = 2.7 for mice.

The baseline model without post-processing performs the worst across all experiments, with high variability among participants. The standard deviation in Jaccard Index and $\mathrm{DSC}$ scores is approximately 20 for rats and 17 for mice, highlighting the inconsistencies in baseline segmentation performance. To mitigate these issues, we tested three different post-processing techniques to refine prompt points and improve segmentation over time.

Applying Local Exploratory Sampling results in a substantial performance boost, improving both datasets by approximately 10 points across all evaluation metrics. This technique consistently ranks second-best overall, outperforming the Kalman Filter in nearly all cases. Interestingly, LES even surpasses Depth-Aware Refinement (DAR) in specific scenarios, particularly when the baseline already provides reasonable prompt points. For instance, in participant p3 (mice), LES achieves a Jaccard Index of 67.9, significantly outperforming DAR at 59.4, with $\mathrm{DSC}$ improving from 64.4 to 73.2. A similar trend is observed in participant p5 (rats) and p6, where LES improves segmentation performance. However, LES struggles when the baseline prompts are highly inaccurate, as it lacks the ability to leverage additional depth information.

The Kalman Filter consistently underperforms compared to both LES and DAR, particularly when the baseline provides poor initial prompts. This is expected, as KF relies heavily on past measurements, which may be unreliable when segmentation starts from a weak initial state. Despite this limitation, Kalman filtering still outperforms the baseline in all cases, demonstrating its ability to smooth predictions and prevent drastic failures. However, its performance lags behind other refinement techniques, particularly when dealing with highly variable or poor-quality input prompts.

The best-performing method on average is Depth Aware Refinement (DAR), which consistently improves results across all participants. This method excels in stabilizing segmentation accuracy, particularly for rats, where it brings all participants to similar performance levels, showcasing its robustness. For instance, participant p2 (rats) sees a remarkable improvement, with Jaccard Index increasing from 16.5 to 61.2 and $\mathrm{DSC}$ jumping from 19.4 to 68.2, representing absolute gains of +44.7 in $\mathrm{J}$ and +48.8 in $\mathrm{DSC}$. However, DAR's effectiveness diminishes when the baseline is already strong, suggesting a potential performance ceiling at around $\mathrm{J}$ = 74 and $\mathrm{DSC}$ = 82. This could be due to inaccuracies in depth estimation, which, while helpful, is not always perfect. For the mice dataset, the improvements are less pronounced. DAR still significantly outperforms all other methods but achieves smaller absolute gains: Jaccard Index improves by +27.4 for rats but only +16.2 for mice, while $\mathrm{DSC}$ increases by +29.8 for rats but only +17.3 for mice. This reduced improvement can be attributed to camera tilting during data collection, leading to imperfect depth estimations, which may impact performance. Nevertheless, DAR remains the most effective post-processing approach, outperforming both LES and KF in nearly all cases.

Among the three tested post-processing techniques, Depth Aware Refinement (DAR) is the most effective overall, demonstrating strong improvements in segmentation accuracy and robustness. Local Exploratory Sampling (LES) performs second-best, particularly excelling when baseline prompts are reasonable. The Kalman filter (KF), while offering improvements over the baseline, struggles in cases with highly unreliable initial prompts. These findings highlight the importance of incorporating depth information and adaptive prompt correction strategies to enhance video segmentation performance.

Table \ref{tab:result} illustrates the impact of combining different post-processing techniques. As a reference, we consider the pipeline using depth aware refinement, which performed best when tested in isolation against other post-processing methods and the baseline. The first key observation is that combining other post-processing techniques without depth aware refinement does not surpass its performance. In fact, it slightly underperforms compared to using LES alone for rats, though it shows improvements for mice. Furthermore, integrating depth aware refinement with other methods consistently enhances performance, though the degree of improvement varies depending on the dataset. For rats, the changes remain modest across different combinations incorporating depth aware refinement (standalone: $\mathrm{J}$ = 66.2, $\mathrm{DSC}$ = 74.2; best combination—DAR + LES: $\mathrm{J}$ = 68.4, $\mathrm{DSC}$ = 75.8), yielding an increase of about 2 points in both metrics. In contrast, for mice, the performance gains are substantial. Notably, combining all three post-processing techniques results in the highest improvement, with $\mathrm{J}$ increasing by 7.3 and $\mathrm{DSC}$ by 7.5.
\section{Conclusion}
We propose a novel pipeline that leverages eye-tracking to generate prompt points, which are then used to produce segmentation masks via a fast zero-shot segmentation model. Our results demonstrate significant variability in performance across different subjects. However, we show that by employing effective post-processing techniques, these performance differences can be mitigated, thereby enhancing the overall applicability of the approach regardless of the subject. For future work, exploring the impact of different hardware setups on segmentation performance could provide further insights and potential optimizations for real-world deployment.

\section{COMPLIANCE WITH ETHICAL STANDARDS}
This study was performed in line with the principles of the
Declaration of Helsinki. Ethical approval was granted by the Interfaculty Ethics Committee of RWTH Aachen University
(reference number 03/25). Ethical approval for animal recording was not required as the videos were obtained without direct interaction or handling of the animals. The recordings were sourced from previous studies, ensuring no additional impact on the animals during this work.

\section{ACKNOWLEDGMENTS}
We gratefully acknowledge the Project Aria \cite{engel2023project} from Meta Platforms, Inc. for donating the glasses used in this work. We gratefully acknowledge the time, effort, and cooperation of
all participants in the subjective experiments.

\bibliographystyle{IEEEtran} % use IEEEtran.bst style
\bibliography{IEEEabrv,bib}

\end{document}